\documentclass[10pt,twocolumn,letterpaper]{article}

\usepackage{iccv}
\usepackage{times}
\usepackage{epsfig}
\usepackage{graphicx}
\usepackage{amsmath}
\usepackage{amssymb}
\usepackage{fitbox}
\usepackage{graphicx}
\usepackage[hypcap=true]{caption}
\usepackage{hyperref}
\usepackage[accsupp]{axessibility}

\iccvfinalcopy 

\ificcvfinal\fi

\begin{document}

\title{Generating Realistic Images from In-the-wild Sounds}

\author{
  Taegyeong Lee~\hspace{0.5cm}
  Jeonghun Kang~\hspace{0.5cm}
  Hyeonyu Kim~\hspace{0.5cm}
  Taehwan Kim \\
  Artificial Intelligence Graduate School, UNIST \\
  \texttt{\small \{taegyeonglee, jhkang, khy0501, taehwankim\}@unist.ac.kr}
}

\twocolumn[{%
\renewcommand\twocolumn[1][]{#1}%
\maketitle
\begin{center}
    \centering
    \includegraphics[width=1\textwidth]{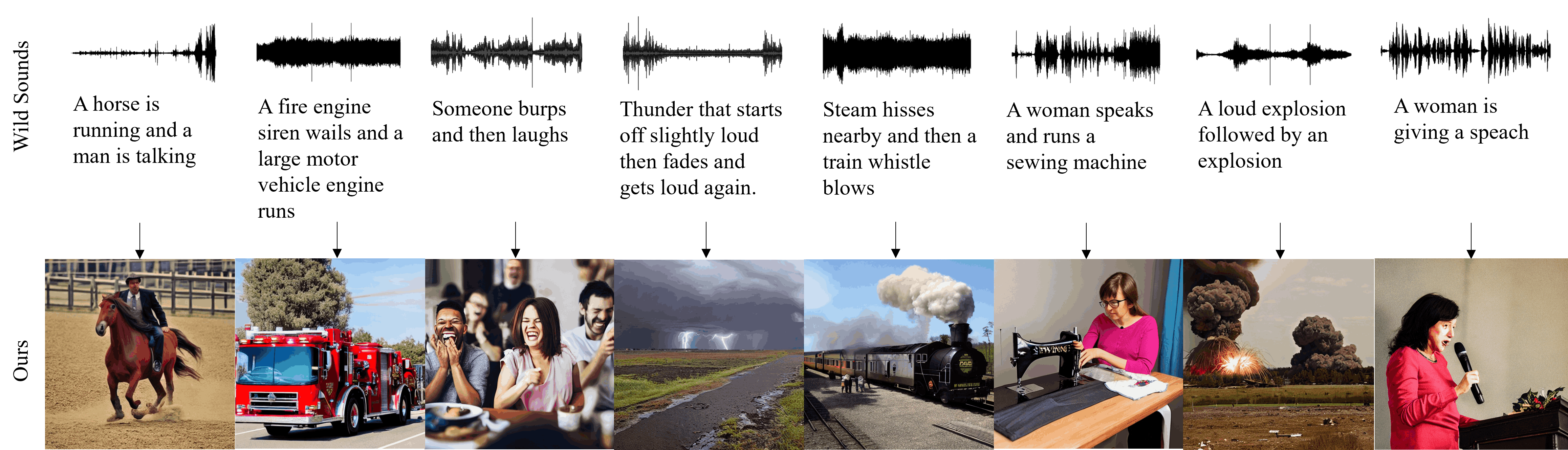}
    \captionof{figure}{\small{\textbf{Generated images from in-the-wild sounds.} We propose an novel approach to generate realistic images from wild sounds. Our model is capable of representing wild sounds as images without large paired datasets of sound and image.}}
    \label{figure:teaser}
\end{center}%
}]

\ificcvfinal\thispagestyle{empty}\fi

\begin{abstract}
Representing wild sounds as images is an important but challenging task due to the lack of paired datasets between sound and images and the significant differences in the characteristics of these two modalities. Previous studies have focused on generating images from sound in limited categories or music. In this paper, we propose a novel approach to generate images from in-the-wild sounds. First, we convert sound into text using audio captioning. Second, we propose audio attention and sentence attention to represent the rich characteristics of sound and visualize the sound. Lastly, we propose a direct sound optimization with CLIPscore and AudioCLIP and generate images with a diffusion-based model. In experiments, it shows that our model is able to generate high quality images from wild sounds and outperforms baselines in both quantitative and qualitative evaluations on wild audio datasets.
\end{abstract}

\section{Introduction}
Sound is one of the most important senses for humans, along with vision. Its dynamic and time-changing characteristics make it a more rich and complex modality than text or image~\cite{lee2022sound}. Therefore, generating corresponding images given sounds has many applications such as background picture generation used in movies and visualization \cite{zhang2018study} and explanation of sound to help people with hearing impairment \cite{azar2007sound}. However, because of differences in modalities and the lack of paired datasets between sound and image, representing wild sounds as images is a challenging task. For these reasons, previous studies~\cite{hao2022attention,lee2020crossing,li2018creating,shim2021s2i,wan2019towards} have generated images within limited sound categories or through music. 
 But generating images in limited sound categories has limitations in representing the wild sounds. It is difficult to represent the complex sounds of multi-domain. Music is composed of elements such as melody and rhythm and contains rich information~\cite{jeong2021traumerai}, but it is fundamentally different from wild sounds that encompass diverse environments and multi-domains. In addition, the quality of images generated from sound by previous studies ~\cite{hao2022attention,shim2021s2i,wan2019towards,wu2022wav2clip} is significantly inferior compared to that of text-guided image generation models ~\cite{ramesh2022hierarchical,rombach2022high,saharia2022photorealistic}.

To solve these problems, we propose a novel approach that uses a pre-trained Audio Captioning Transformer (ACT) ~\cite{mei2021audio} and a diffusion-based model called Stable Diffusion~\cite{rombach2022high}. Unlike previous studies \cite{ hao2022attention,lee2022sound, wan2019towards} that attempted to generate images from sound by mapping it into limited categories such as dogs or humans, we describe sounds in greater detail by converting them into audio captions by using the ACT \cite{mei2021audio} model. Then, we generate images using a pre-trained Stable Diffusion \cite{rombach2022high} model. This approach address the differences in modalities and enables high-quality image generation from a sound without requiring large paired training datasets between sounds and images.

 However, since our purpose is to represent the rich and dynamic characteristics of wild sounds in images, simply generating images from audio captions is not sufficient. Therefore, we propose a novel approach to address this issue.  Firstly, we propose to exploit \emph{audio attention}. Audio attention is a value of probability used by the ACT \cite{mei2021audio} model to generate an audio caption. We first use this audio attention to represent the rich and dynamic characteristic of sound as an image. Secondly, we introduce \emph{sentence attention} to emphasize objects from the generated audio caption. To visualize sound, it is important to consider not only the characteristics of the sound, but also to emphasize the objects in the sound within it.

Furthermore, we propose \emph{direct sound optimization} to optimize images further for corresponding sounds. We generate a latent vector from audio caption with CLIP \cite{radford2021learning} text encoder, and initialize it with audio attention and sentence attention to get a sound optimized latent vector. 
After that, we generate an image with Stable Diffusion using the latent vector as a conditioning vector, and optimize it through AudioCLIP \cite{guzhov2022audioclip} similarity and CLIPscore \cite{hessel2021clipscore}. Through this process, we address the modality gap between audio and image and are able to tackle the challenging problem of generating realistic and dynamic images from wild sounds.

In summary, our contributions are as follows: 
\begin{itemize}
    \item We propose a novel approach that uses audio captioning and diffusion based text-to-image model to generate a high quality image without large paired datasets between sounds and images.
     \item We propose audio attention and sentence attention to generate images that represent the characteristics of multi-domain and time-varying dynamic sounds. In addition, we introduce direct sound optimization with CLIPscore and AudioCLIP similarity for further enhancement.
      \item In experimental results, our model is able to generate faithful and high quality images from in-the-wild input sounds and outperforms baselines in both quantitative and qualitative evaluations.
\end{itemize}

\section{Related work}
\textbf{Text-guided Image Generation} Text-gudied image generation has been widely covered along with the progress in image synthesis. Previous works \cite{crowson2022vqgan, kim2022verse, nichol2021glide, ramesh2022hierarchical, ramesh2021zero, rombach2022high, saharia2022photorealistic} used diffusion \cite{ho2020denoising} model, transformer \cite{ child2019generating,vaswani2017attention}, and VQ-VAE \cite{esser2021taming,van2017neural} to synthesize high quality images and adopted ViT \cite{dosovitskiy2020image}, T5 \cite{raffel2020exploring}, and mostly CLIP \cite{radford2021learning} for the understanding of the high-dimensional text and image data. In our work, we tackle audio conditional image synthesis with stable diffusion \cite{rombach2022high} as it can handle diverse conditioning inputs and is also easily accessible.

\textbf{Sound-guided Image Generation} There have been many studies \cite{hao2022attention,lee2020crossing,li2018creating,shim2021s2i,wan2019towards, yang2020diverse} that generate images from sounds. Wan \etal~\cite{wan2019towards} used a conditional GAN model to generate images from sounds for 9 categories, with the image resolution of 64$\times$64. They used SoundNet \cite{aytar2016soundnet} for the sound dataset and crawled images from the internet to create a paired dataset of 10,701 sound-image pairs for training. Wav2clip~\cite{wu2022wav2clip} extends CLIP~\cite{radford2021learning} in the audio dimension and performs contrastive learning using audio, image, and text. The model represents audio features from audio, image and text and use the results as input to VQGAN-CLIP~\cite{crowson2022vqgan} to generate images that visualize the audio. Other past studies~\cite{hao2022attention,lee2020crossing,shim2021s2i} generate images from bird sounds and music. One of the representative examples of generating visual information from music information is TräumerAI~\cite{jeong2021traumerai}. TräumerAI analyzes music, extracts features, maps them onto the feature space of visual information, and generates videos using StyleGAN~\cite{karras2019style} model. However, these have limitations in generating images from wild sounds because the generated images are limited to specific categories. To address this problem, we propose a novel approach with audio attention, sentence attention, and direct sound optimization, and, to the best of our knowledge, we are first to use the diffusion-based model to generate high quality images from wild sounds.

\textbf{Sound-guided Image Manipulation}
A study~\cite{lee2022sound} was also conducted to map sound, text, and image information into a single space by extending the CLIP~\cite{radford2021learning} model which mapped text and image information to a single feature space. In the proposed method, a latent vector based on a sound feature is generated and fed into the StyleGAN2~\cite{viazovetskyi2020stylegan2} model to generate an image. While this method utilizes the semantic information of sound, single domain mapping makes it difficult to manipulate images generated from wild sounds. 

\begin{figure*}[h] 
\begin{center}
\includegraphics[width=1\textwidth]{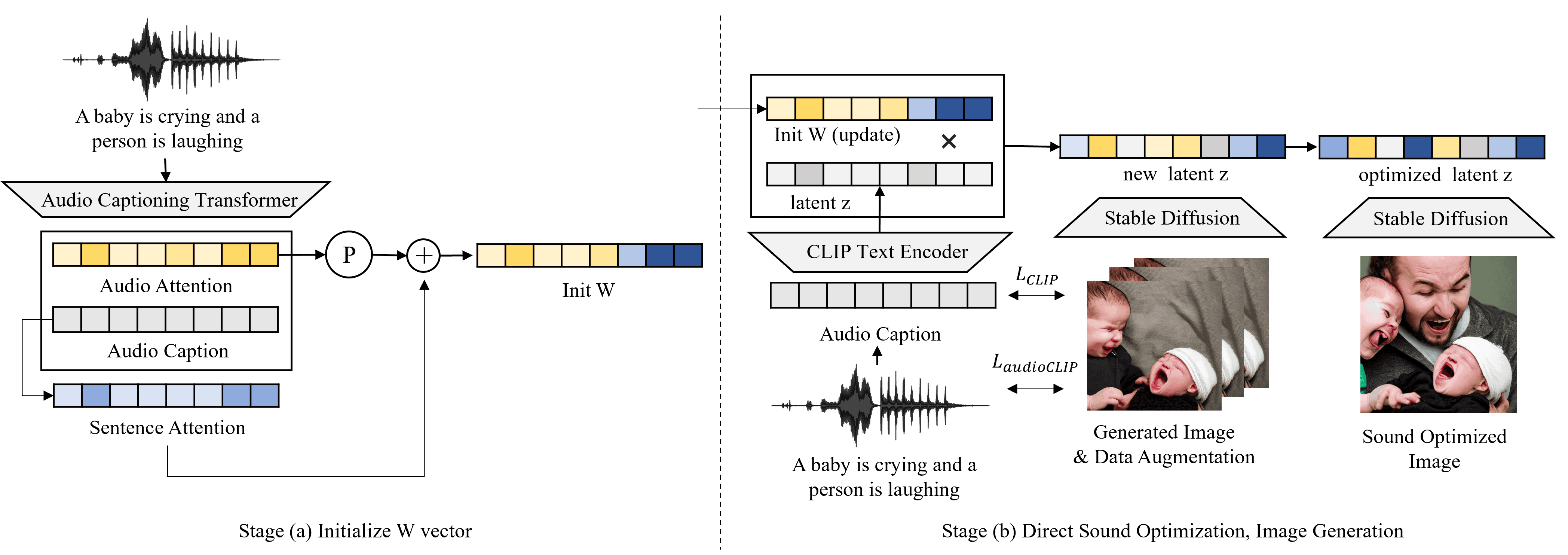}
\end{center}
\caption{\textbf{Overall architecture of our model}. Our model consists of two stages. Stage (a) is the step of initializing the $W$ vector with attentions, and Stage (b) is direct sound optimization and image generation process.}
\label{figure_overview}
\end{figure*} 

\section{Method}
In this section, we introduce our novel approach to represent wild sounds as images. Our model consists of two stages as shown in Figure~\ref{figure_overview}. 
In the first stage, Figure~\ref{figure_overview} (a), we initialize $W_{init}$ from input audio using Audio Captioning Transformer (ACT)~\cite{mei2021audio}, involving audio caption, audio attention and sentence attention. In the second stage, Figure~\ref{figure_overview} (b), we generate images with Stable Diffusion~\cite{rombach2022high} by inputting $W_{init}$ and text latent $z$, and optimize $W_{init}$ for realistic sound characteristics.

\subsection{Audio Captioning}
Previous studies \cite{hao2022attention,lee2022sound,wan2019towards} have generated images by mapping sounds into limited categories. However, these approaches may result in the loss of the rich characteristics of the sound; it will map the sound of "a person walking on a beach with waves" to an image of either waves or a person. To address this issue, we transform sound into audio caption rather than mapping them into a limited category so that one gets more information from sound than from a single label. In this paper, we use a pre-trained ACT \cite{mei2021audio} model for audio captioning, which has a vanilla transformer~\cite{vaswani2017attention} structure and was trained on the Audiocaps~\cite{kim2019audiocaps} dataset. Initial weights of the encoder are initialized with the weights of pre-trained Deit~\cite{touvron2021training} and the encoder was pre-trained with Audioset~\cite{gemmeke2017audio} audio tagging task. By generating audio caption from sound, we use richer information for visualization, such as multi-domain and actions, that cannot be provided with single labels.

\subsection{Attentions and Positional Encoding}
\textbf{Audio Attention} The generated audio captions have more information than single labels, but may not be able to represent the rich intensity and dynamic characteristics of sound. For example, "laughing loudly" is different from "laughing a little", and "thunder from afar" and "thunder from close up" are different. We propose a novel approach, called audio attention, which can effectively represent these features of sound. We use the value of probability, used by the decoder of the ACT model to generate the audio caption as audio attention. It is unique to each sound and allows better representation of the sound than audio captions.  For example, if the sound of thunder is given as an input, the audio caption would generate "Thunder is striking very close". As the sound of thunder gets louder, the value of "Thunder" in the audio attention would also increase accordingly. Through this process, we are able to effectively represent the rich and dynamic characteristics of sound through images.

\textbf{Sentence Attention} We generate audio captions from sound and use audio attention to represent sound characteristics as images. However, in order to visualize sound, it is important not only to represent the characteristics of sound, but also to emphasize the object in the sound. For example, if the input sound is the sound of a person laughing loudly, audio attention will focus more on the laughter sound than on the person. However, since we need to represent the object 'person' as an image, we require additional approach. We propose sentence attention that emphasizes nouns in the audio caption. It is a probability that a word in the sentence is a noun. In this paper, we use FLAIR~\cite{akbik2018coling} model to calculate the probability that each word in the audio caption is a noun. By using sentence attention, we can emphasize the object in the sound, which greatly facilitates the visualization of sound.

\textbf{Positional Encoding}
We empirically find that it is beneficial to use positional encoding to optimize attentions to the text embedding space. After trials, we use Equation~\ref{eq1} to compute the positional encoding $P$ of audio attention.

\begin{equation}\label{eq1}
P = \frac{1}{2 + e^{2-0.5x}}
\end{equation}

\subsection{Initializing $W_{init}$ vector}
We generate an $W_{init}$ vector with audio attention, sentence attention and positional encoding. First, we apply positional encoding to audio attention. Second, add sentence attention and audio attention. In this process, in order to emphasize the object, we multiply the noun of audio attention by a weight $\lambda_{a}$ to relatively increase the weight of the sentence attention (we set $\lambda_{a}$ = 0.1 after experiments). Finally, after tokenizing audio caption, reshape $W_{init}$ to the shape of audio caption token. Figure~\ref{figure_overview}~(a) shows the process of initializing $W_{init}$ vector along with Equations~ \ref{eq_w1}, \ref{eq_w2}, \ref{eq_w3} where $t$ is the audio caption, $S$ is a function that extracts sentence attention from audio caption $t$, $P$ is the positional encoding function, $a$ is the audio given as input, $A$ is a function to extract audio attention from audio $a$, $W_n$ is for noun and $W_{wn}$ is for non-noun. We use the $W_{init}$ vector to generate a realistic image that represents the dynamic characteristics of sound.

\begin{equation}\label{eq_w1}
W_{n} = S(t) + \lambda_{a} A(a)_n + P(A(a)_n) 
\end{equation}
\vspace{-0.1cm}
\begin{equation}\label{eq_w2}
W_{wn} = S(t) + A(a)_{wn} + P(A(a)_{wn}) 
\end{equation}
\vspace{-0.1cm}
\begin{equation}\label{eq_w3}
W_{init} = Reshape(Stack(W_{n}, W_{wn}))
\end{equation}

\subsection{Image Generation}
 Let $G$ be the image generation model that takes text embedding as an input. We use one of the current state-of-the-art diffusion models called Stable Diffusion \cite{rombach2022high} as $G$. We use its pre-trained model with LAION-5B~\cite{schuhmann2022laion} to generate real images optimized for open domain sound. We encode the audio caption into latent vector $z$ in the text embedding space using the CLIP \cite{radford2021learning} text encoder. After initializing the $W_{init}$ vector, we multiply the latent vector $z$ to get a new latent vector $z_{n}$. The model generates an image $I_{init}$ with $z_{n}$ as input:

\begin{equation}\label{eq_w5}
I_{init} = G(W_{init} \odot z)
\end{equation}
where $\odot$ is element-wise multiplication.

\subsection{Direct Sound Optimization}
We generate an image by inputting a new latent $z_{n}$ into Stable Diffusion~\cite{rombach2022high}. We are inspired by StyleCLIP~\cite{patashnik2021styleclip} and perform direct sound optimization using CLIPscore~\cite{hessel2021clipscore} between image and audio caption and AudioCLIP~\cite{guzhov2022audioclip} similarity between audio and image. Our total loss is as follows:
\begin{equation}\label{eq2}
L_{total} = \lambda_{aCLIP} L_{aCLIP} \\
+ \lambda_{CLIP} L_{CLIP} 
+ \lambda_{L2} L_{L2}
\end{equation}

\begin{figure}[t] 
\begin{center}
\includegraphics[width=0.95\linewidth]{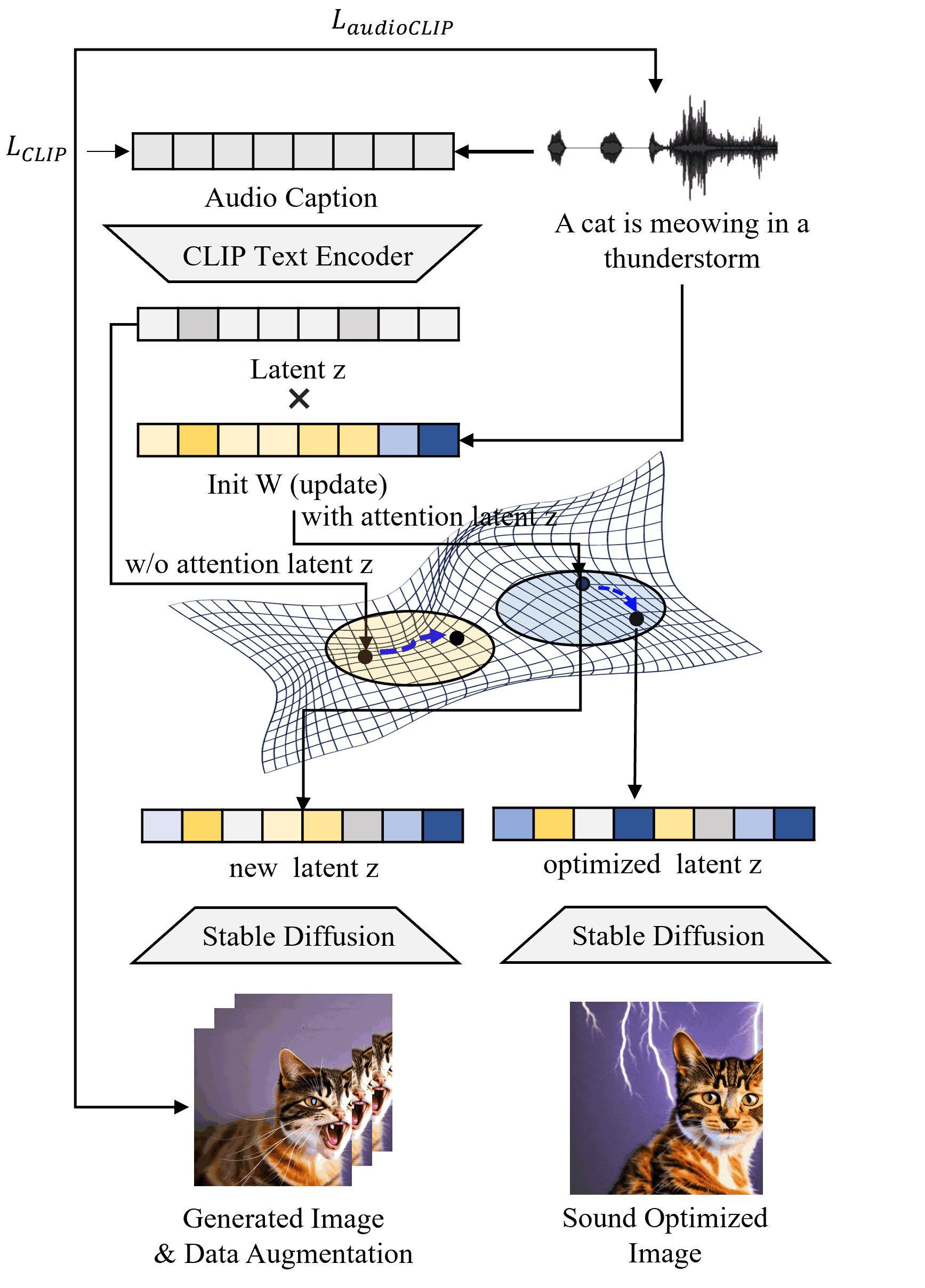}
\end{center}
\caption{\textbf{Process of direct sound optimization.} When initializing latent $z$ with attentions, we optimize in the blue area, whereas without using attention, we optimize in the yellow area. To solve the problem of local minimum and to represent the rich features of sound, attentions are necessary.}
\vspace{-0.3cm}
\label{figure_optimization}
\end{figure}

$L_{aCLIP}$ is the term of similarity between the input audio $a$ and the generated image $I$. $L_{CLIP}$ indicates the similarity between the generated image $I$ and the audio caption $t$. We maximize these two losses. $L_{L2}$ is the L2 norm of the latent vector $z$ generated by the audio caption $t$ and the new latent vector $z_{n}$. Additionally, we augment the image generated by the new latent vector $z_{n}$ with VQGAN-CLIP~\cite{crowson2022vqgan} data augmentation method. In order to focus on the nouns part of the audio when generating the image, we divide $W_{init}$ into nouns $W_{n}$ and without nouns $W_{wn}$. In the direct sound optimization step, the $W_{n}$ is optimized with a higher learning rate because of local minimization problem. By using this, we get more natural results from audio captions and audio. Equation~\ref{eq2} is the loss used in direct sound optimization and Figure~\ref{figure_optimization} shows this process.

\begin{table*}[t]
\vspace*{-\baselineskip}
\noindent
\begin{minipage}{0.7\textwidth}
\centering
\small
\begin{tabular}{lccccccc}
\hline
                              & \multicolumn{7}{c}{Dataset}                                                                                               \\ \cline{2-8} 
                              & \multicolumn{2}{c}{Audiocaps}      & \multicolumn{2}{c}{Clotho}         & Urban8K & \multicolumn{2}{c}{Multi-ESC50} \\
Model                         & CLIPscore$\uparrow$ & IS$\uparrow$ & CLIPscore$\uparrow$ & IS$\uparrow$ & IS$\uparrow$  & IS$\uparrow$     & Yolo$\uparrow$      \\ \hline
Wan~\etal & 0.5298              & 2.52         & 0.5411              & 2.35         & 2.05          & 1.95             & 0.00       \\
W2c-vqgan                     & 0.5249              & 6.69         & 0.4649              & 6.71         & 4.91          & 5.55             & 0.07       \\
Ours                          & \textbf{0.6580}              & \textbf{17.03}        & \textbf{0.5875}              & \textbf{17.47}        & \textbf{6.30}          & \textbf{6.51}             & \textbf{0.7857}       \\ \hline
\end{tabular}
\caption{\textbf{Quantitative evaluations}. We compare our method with baselines on wild \\
audio datasets. IS is Inception score~\cite{salimans2016improved}. Yolo is YOLO score.}
\label{table:baseline}

\end{minipage}%
\begin{minipage}{0.3\textwidth}
\vspace{0.6cm}
\centering
\footnotesize
\begin{tabular}{lcc}
\hline
\multicolumn{1}{c}{} & Wan \etal & W2c-vqgan \\ \hline
Audiocaps            & 95.10\%     & 90.26\%       \\
Clotho               & 96.00\%     & 93.46\%       \\
Urban8K                & 94.70\%     & 87.90\%       \\
Multi-ESC50                & 96.00\%     & 94.80\%       \\
\hline
\end{tabular}
\caption{\textbf{Pairwise comparison with baseline models in human evaluation.} Each cell lists the percentage where our result is preferred over the baseline models.}
\label{table:baseline_human}
\end{minipage}
\end{table*}

\section{Experiments}
\subsection{Datasets}
To effectively evaluate image generation performance from wild sounds, we  use audio captioning datasets \cite{drossos2020clotho,kim2019audiocaps} in a sound-guided image generation task for the first time. The datasets used in the experiments are as follows:\\
 \textbf{Audiocaps}~\cite{kim2019audiocaps} is a paired dataset between sound in the wild and text that labels about 50k of audio data in Audioset~\cite{gemmeke2017audio}. The dataset contains human-annotated captions and various multi-domain sounds. We use the test set consisting of a total of 957 audio files for both qualitative and quantitative evaluations.\\
 \textbf{Clotho}~\cite{drossos2020clotho} is a dataset for audio captioning consisting of 4,981 wild audio files of 15 to 30 seconds duration and 24,905 captions. We use the test set consisting of a total of 1,045 audio files for both qualitative and quantitative evaluations.\\
 \textbf{Urbansound 8K}~\cite{salamon2014dataset} is a dataset for sound classification, consisting of 8,732 audios in 8 classes. To visualize sounds, we exclude abstract classes, and we use 400 audios from a total of 4 classes (dog barking, children playing, car horn, and gun shot) for both qualitative and quantitative evaluation. \\
\textbf{Multi-ESC50}
Wild sounds may contain multi-domain and have domains that change over time. We composed 700 audio files by concatenating two classes
of sounds from the ESC50~\cite{piczak2015esc} dataset, excluding abstract classes. It consists of a total of 7 classes. Three classes have two objects, and four classes consist of object and background. Each class consists of a total of 100 sounds. For example, we combined baby cry class and laughing class.

\subsection{Baselines}
We use Wan \etal~\cite{wan2019towards}, which generates images from sound, and W2-vqgan, which generates open-domain images with VQGAN~\cite{crowson2022vqgan} through the representing sound features proposed in Wav2CLIP~\cite{wu2022wav2clip}, as baselines. We train the Wan \etal model for 9 classes with paired datasets between sounds and images on SoundNet \cite{aytar2016soundnet}, as described in their paper. This was because paired datasets between sounds and images in open domains are difficult to obtain. The W2c-vqgan is a model that uses Wav2CLIP and VQGAN \cite{crowson2022vqgan} pre-trained with ImageNet \cite{deng2009imagenet}.

\begin{figure*}[t] 
\includegraphics[width=1.0\textwidth]{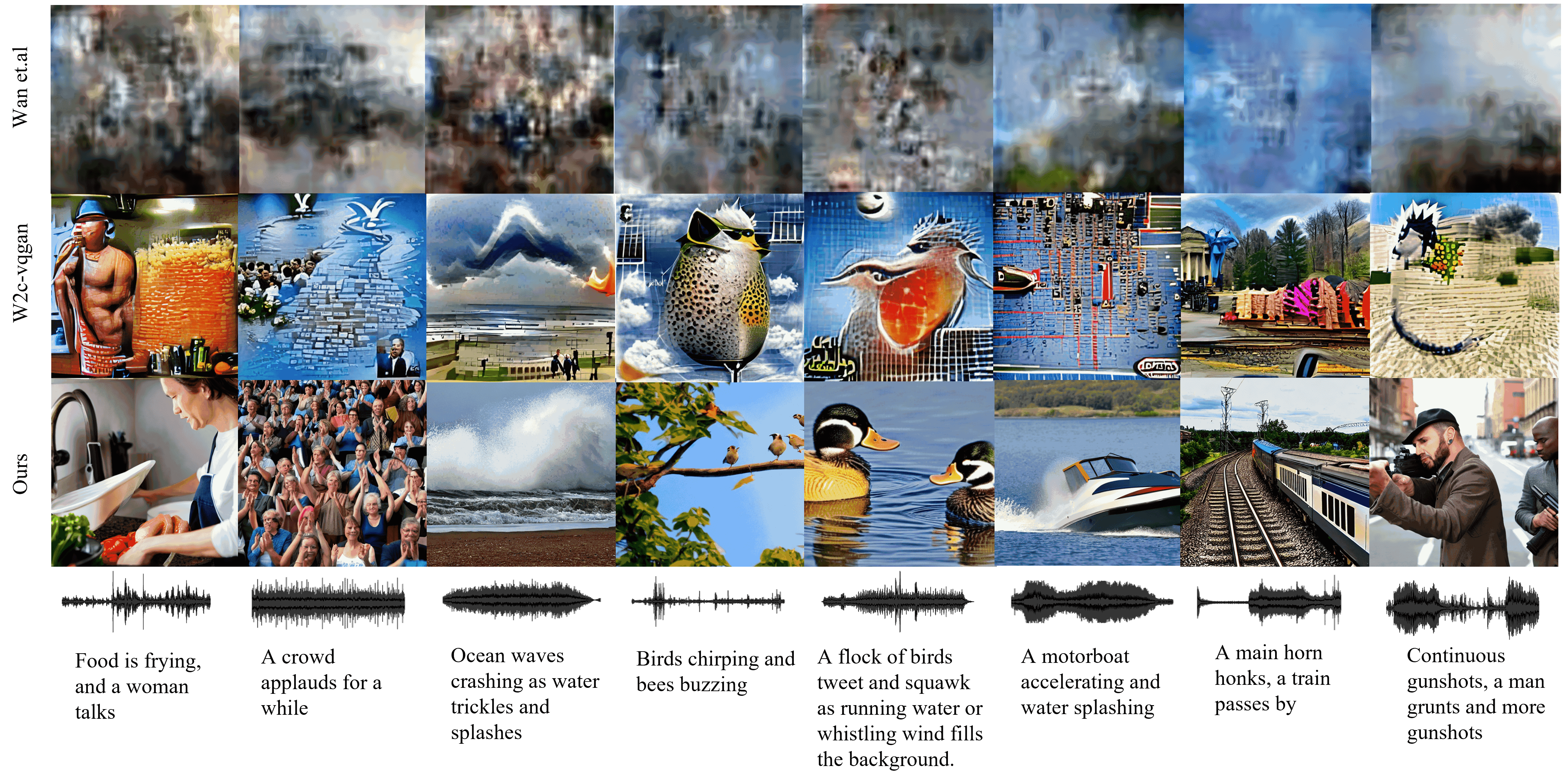}
    \caption{\textbf{Comparison of sound-guided image generation result.} Given audio inputs, we generate images using our models and baselines~\cite{wan2019towards,wu2022wav2clip}. Text is ground-truth (GT) audio caption of audio datasets~\cite{drossos2020clotho, kim2019audiocaps}. Generated samples here are randomly chosen.}
\label{figure_comparison}
\vspace{-0.4cm}
\end{figure*}

\subsection{Evaluation Metrics}
There is no metric available to directly evaluate the similarity between sound and image. Therefore, we propose to evaluate the performance of sound-to-image synthesis models indirectly using CLIPscore~\cite{hessel2021clipscore}, Inception score~\cite{salimans2016improved} and YOLO score:\\
\textbf{CLIPscore}~\cite{hessel2021clipscore} is a metric that evaluates the similarity between text and images. In this paper, we calculate the score between the audio captions in ground-truth of each dataset~\cite{drossos2020clotho,kim2019audiocaps} and the generated images from wild sounds. In the case of UrbanSound8K and Multi-ESC50, CLIPscore could not be calculated because there was no ground truth audio caption available.\\
\textbf{Inception score}~\cite{salimans2016improved} is a metric used to evaluate the quality and diversity of the generated images. \\
\textbf{YOLO score} To evaluate whether there is an object in the image that matches the sound or not, we use an object detection model Yolo v5~\cite{redmon2016you}.

\subsection{Implementation Details}
We used a single NVIDIA A100 80G for our experiments, and performed 10 steps of direct sound optimization using the Adam~\cite{kingma2014adam} optimizer with $\lambda_{aCLIP}$ = 0.9, $\lambda_{CLIP}$ = 1.0, $\lambda_{L2}$ = 0.01 and set the learning rate for $W_{n}$ to 0.01 and learning rate for  $W_{wn}$ to 0.001. We used a PLMS sampler and we performed the DDIM step 40 times. Our model and W2c-vqgan have an image resolution of 512$\times$512, while Wan \etal~\cite{wan2019towards} has an image resolution of 64$\times$64. More detail can be found in the supplementary material.

\subsection{Comparisons with the baselines}\label{baselines}
\subsubsection{Quantitative Evaluation}
We quantitatively evaluate the similarity between sound and generated images on the four datasets. We compare the performance of our model with baseline models, Wan \etal~\cite{wan2019towards} and W2c-vqgan \cite{wu2022wav2clip}. As shown in Table~\ref{table:baseline}, our model shows a significant CLIPscore \cite{hessel2021clipscore} improvement compared to the baseline models on the wild sound datasets (Audiocaps~\cite{kim2019audiocaps} and Clotho~\cite{drossos2020clotho}). In the Yolo score evaluation of object matching between the given sounds and generated images, while the baseline showed a very low score, our model achieved 78.57\%, outperforming the baseline by a significant margin. Also, our model outperforms the baseline models by a large margin in Inception score~\cite{salimans2016improved} on all datasets. These results demonstrate that our model generates high-quality images optimized for sound, surpassing baseline models on all datasets. It is also noteworthy that evaluations on Clotho,  Urbansound 8K, and Multi-ESC50 are zero-shot and thus they show the strong performance of our model for even zero-shot setting.

\subsubsection{Qualitative Analysis}
\vspace{-0.2cm}
Because the automatic metric cannot comprehensively evaluate the quality of generated images from sounds, we conducted a side-by-side human evaluation study on Amazon Mechanical Turk. We use a total of 300 audio files from three datasets, randomly sampling 100 audio files from each dataset. We recruit 30 participants for each study.
In each study, participants listened to a sound and were asked to select one of two images generated by different models. Participants answer the following questionnaire: Listen to the following audio and choose the most appropriate image. We randomly swap the images generated from baseline and our model to remove bias. As shown in Table~\ref{table:baseline_human}, the percentage of participants who chose our model to be more suitable when comparing between our model and Wan~\etal~\cite{wan2019towards} is 95.10\% in Audiocaps, 96.00\% in Clotho, 94.70\% in Urban8K and 96.00\% in Multi-ESC50. 
When comparing our model with W2c-vqgan~\cite{wu2022wav2clip}, the results show a similar trend to Wan~\etal, and the participants select W2c-vqgan model slightly higher than Wan~\etal when compared to our model. In all experiments, a large portion of participants chose generated image by our model as the best. In addition, as shown in Figure~\ref{figure:teaser},~\ref{figure_comparison}, our model generates realistic images of detailed objects and scenes but W2c-vqgan and Wan \etal generate abstract images on wild audio datasets. Specifically, W2c-vqgan generates abstract images of birds, bees, and the sky when sound of ``birds chirping and bee buzzing" are given as input, whereas the images generated by Wan \etal are difficult to interpret in terms of the correspondence between sound and image. These results indicate that our model better represents sound as images compared to the baseline models and demonstrate a significant improvement in image quality performance. More generated samples of our model and baselines are included in supplementary material.

\vspace{-0.25cm}
\subsection{Impact of Magnitude of the Sound}
\begin{figure}[t] \centering
\includegraphics[width=0.48\textwidth]{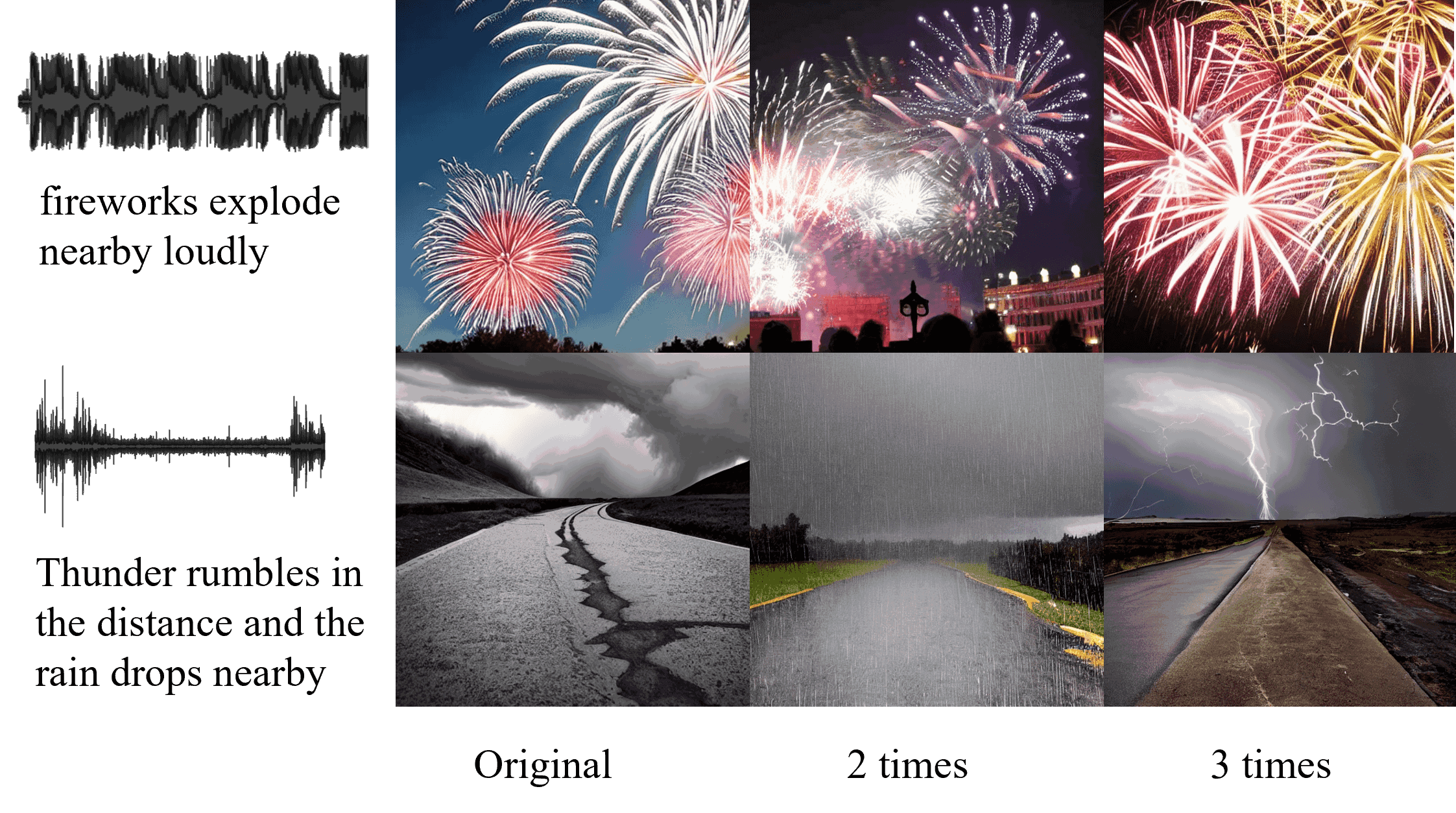}
    \caption{\textbf{Generated images by different amplitudes of sounds.} 2 times means the sound amplitude is doubled, and 3 times means it is tripled compared to the original sounds. Text is ground-truth audio caption of audio datasets~\cite{drossos2020clotho, kim2019audiocaps}.} \label{figure_impact}
\end{figure}

Altering images based on the magnitude of sound is one of the methods to demonstrate the rich characteristics of sound in a sound-to-image task~\cite{wan2019towards}. We also investigate the impact of sound amplitude on the resulting image by amplifying the sound two and three times. As shown in Figure~\ref{figure_impact}, we observe that the images became more exaggerated as the sound become louder. These results demonstrate that our model generates richer and more relevant images depending on the sound.

\subsection{Expansion of Sound Modality}
\begin{figure}[t] \centering
\includegraphics[width=0.48\textwidth]{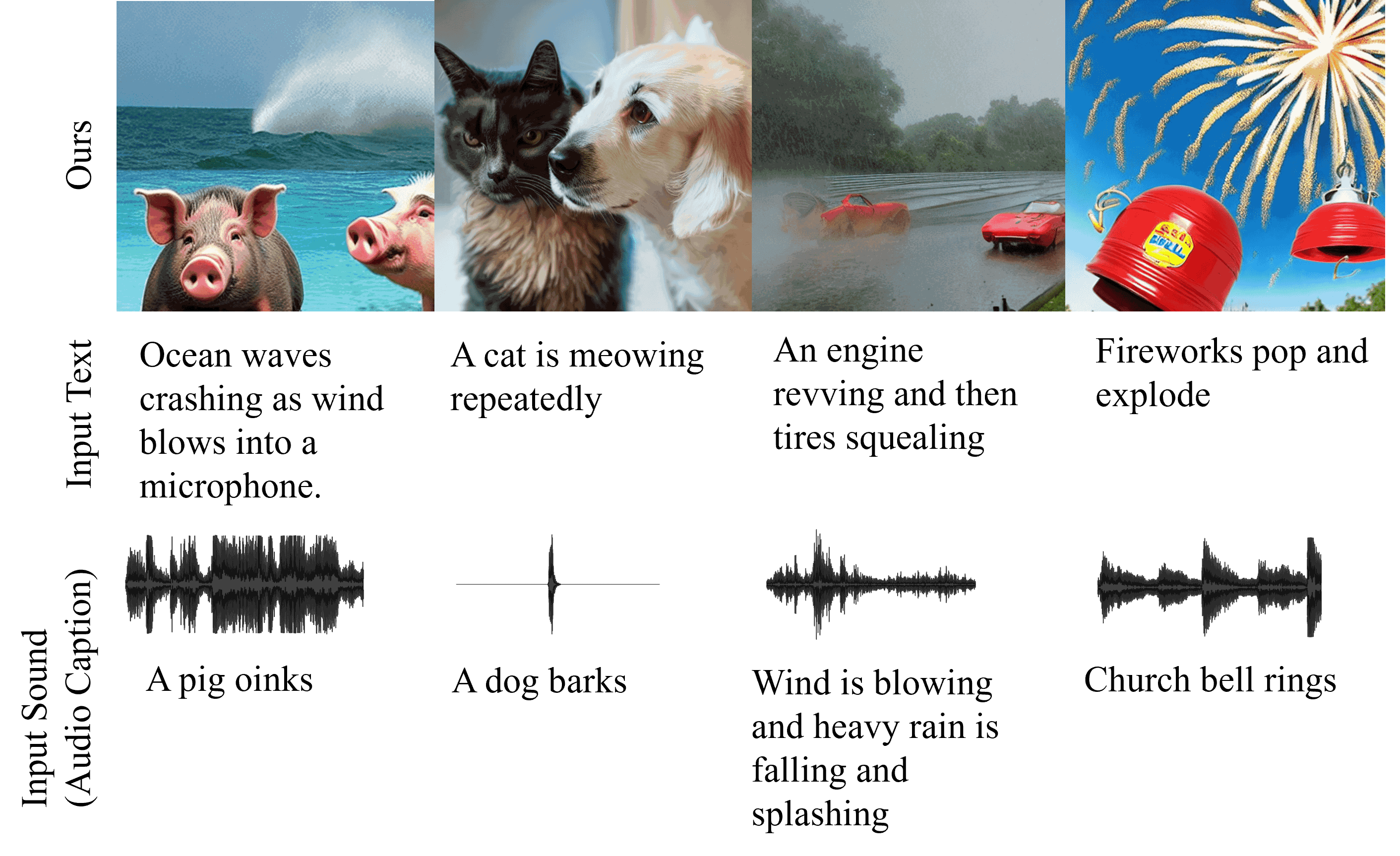}
\caption{\textbf{Generated images by combining sound and text modality}. We generate images by using text and sound as inputs.} 
\vspace{-0.2cm}
\label{figure_expand}
\end{figure}

Our model can expand the sound modality by combining it with text modality, as we use a pre-trained text-guided image generation~\cite{rombach2022high} model. As shown in Figure~\ref{figure_expand}, we observe the generation of an image describing ``a car driving in the rain" when inputting the text ``An engine revving and then tires squealing" and sound of ``Wind is blowing and heavy rain is falling and splashing.". This allows us to leverage the advantages of both rich sound modality and the ability of text modality to specify objects concretely.

\section{Ablation Study}
We provide ablations with respect to our model on Audiocaps~\cite{kim2019audiocaps} and Multi-ESC50 with wild sounds and multi-object sounds. We conducted human evaluation to evaluate the similarity between sound and the generated images from a human perspective. We compared our model with 5 ablation models, including a model without attentions and direct sound optimization (w/o all), a model without all attentions (w/o attns), a model without audio attentions (w/o a attn), a model without sentence attentions (w/o s attn), and a model without direct sound optimization (w/o opt). All experimental environments are the same as in Section \ref{baselines}.

\begin{figure}[t] \centering
\includegraphics[width=0.47\textwidth]{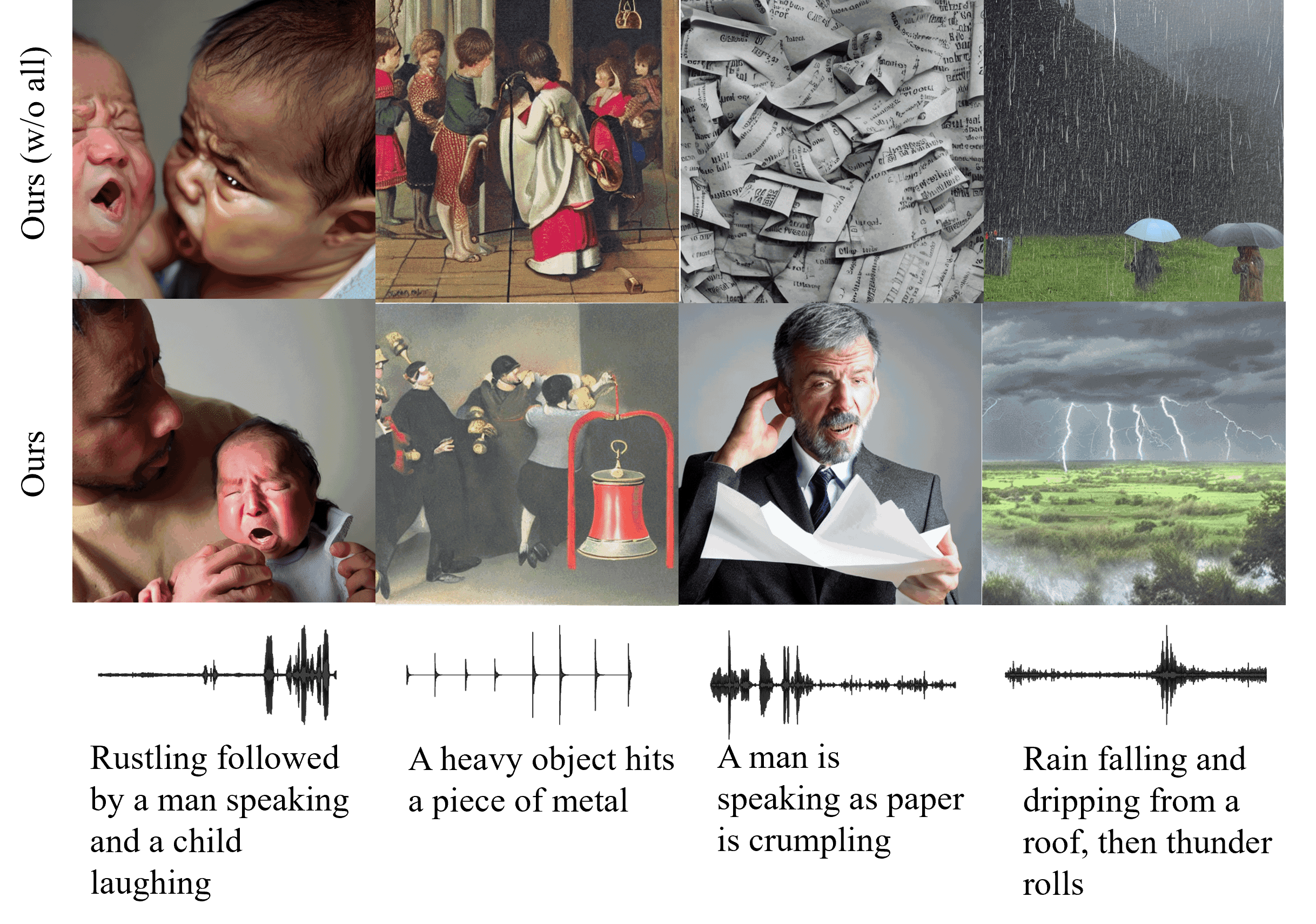}
    \caption{\textbf{Ablation study on all attentions and direct sound optimization.} Ours(w/o all) is a model that does not have both attentions and direct sound optimization. Text is ground-truth audio caption of audio datasets \cite{kim2019audiocaps}.} 
\label{figure_ablation_01}
\end{figure}

\textbf{Ablation: Attentions and Optimization}
We compare our model and  w/o all model. W/o all model does not have attentions and direct sound optimization and generates images by inputting simply audio captions generated by the ACT \cite{mei2021audio} model into Stable Diffusion \cite{rombach2022high}. Table~\ref{table:ablation} shows that our model achieves better performance than the w/o all model. Furthermore, on the Multi-ESC50 dataset, the Yolo score for the w/o all model was 41.86\%, showing a significant difference compared to our model's score of 78.57\%. In the human evaluation, as shown in Table~\ref{table:ablation_human}, the percentage of participants who chose our model to be more suitable between our model and w/o all is 63.46\% in Audiocaps~\cite{kim2019audiocaps} and 70.30\% in Multi-ESC50. As shown in Figure~\ref{figure_ablation_01}, when given the input of the sound of ``a man speaks as paper crumples", our model generates both a man and paper, while w/o all model does not generate the man. These results indicate that our model generates higher-quality sound-optimized images in the eyes of human observers and represent more faithful to the input sounds compared to the model without all.

\begin{table}[t]
\centering
\small
\begin{tabular}{lcccc}
\hline
                     & \multicolumn{4}{c}{Dataset}                                          \\ \cline{2-5} 
\multicolumn{1}{c}{} & \multicolumn{2}{c}{Audiocaps}      & \multicolumn{2}{c}{Multi-ESC50} \\
Model                & CLIPscore$\uparrow$ & IS$\uparrow$ & IS$\uparrow$     & Yolo$\uparrow$      \\ \hline
w/o all              & 0.6538              & 16.13        & 5.80             & 0.4186       \\
w/o attns            & 0.6563              & 16.57        & 6.24             & 0.6914       \\
w/o s attn           & 0.6573              & 16.63        & 6.32             & 0.7285       \\
w/o a attn           & 0.6564              & 16.75        & 6.46             & 0.7243       \\
w/o opt              & \textbf{0.6581}              & 16.00        & 6.15             & 0.7200       \\ \hline
Ours                 & 0.6580              & \textbf{17.03}        & \textbf{6.51}             & \textbf{0.7857}       \\ \hline
\end{tabular}
\caption{\textbf{Quantitative evaluations}. We compare our method with ablation models on wild
audio datasets. IS is Inception score~\cite{salimans2016improved}. Yolo is YOLO score.}
\label{table:ablation}
\end{table}

\begin{table}[t]
\centering
\small
\begin{tabular}{lcc}
\hline
           & Audiocaps & Multi-ESC50 \\ \hline
w/o all    & 63.46\%   & 70.30\%     \\
w/o attns  & 60.93\%   & 64.60\%     \\
w/o a attn & 59.20\%   & 60.10\%     \\
w/o s attn & 58.40\%   & 58.30\%     \\
w/o opt    & 64.60\%   & 62.00\%     \\ \hline
\end{tabular}
\caption{\textbf{Pairwise comparison with ablation models in human evaluation.} w/o attns is model without all attentions. w/o a attn is model without audio attention. w/o s attn is model without sentence attention. Each cell lists the percentage where our result is preferred over the baseline models.}
\vspace{-0.6cm}
\label{table:ablation_human}
\end{table}

\begin{figure}[t] \centering
\includegraphics[width=0.47\textwidth]{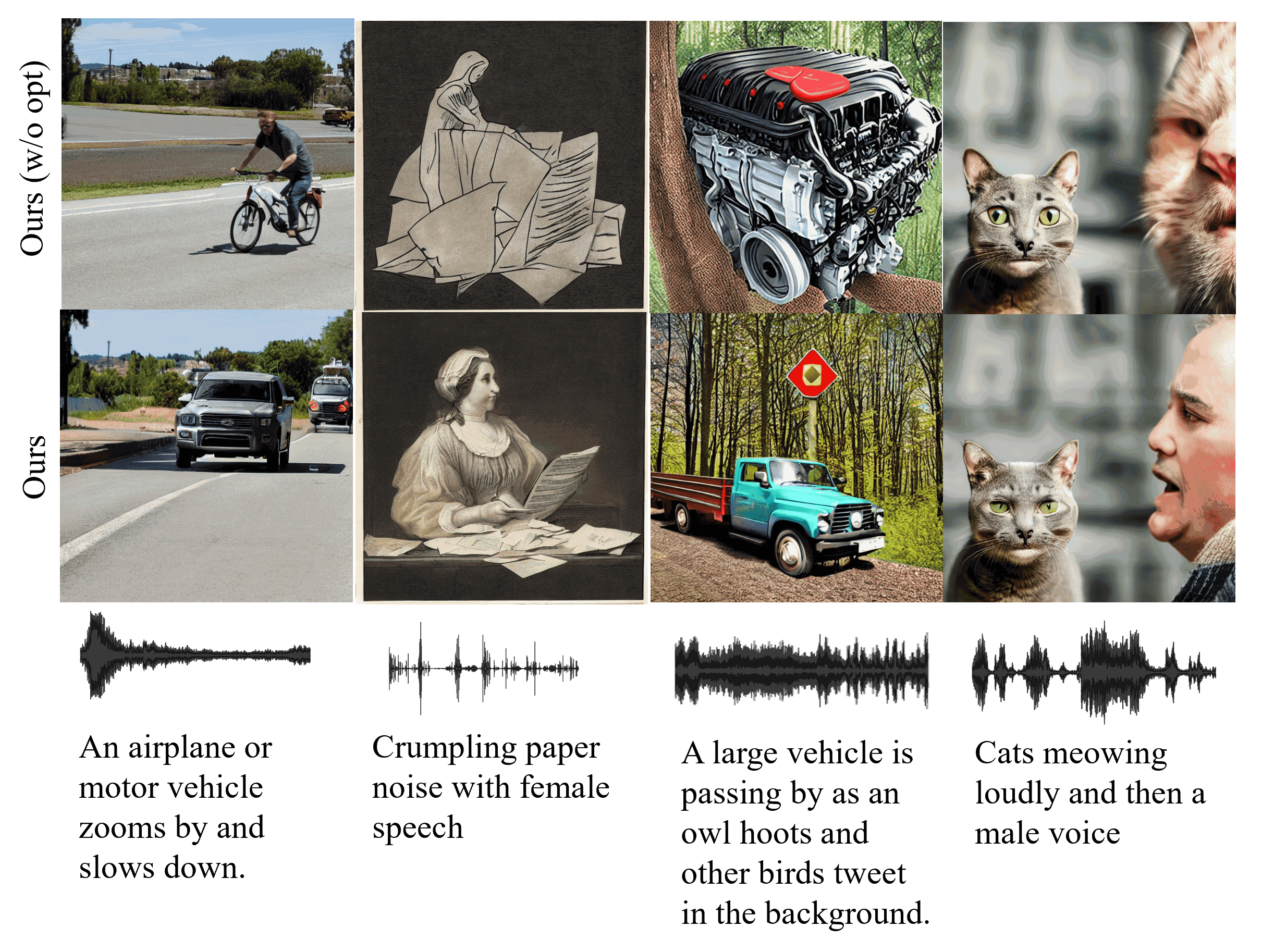}
    \caption{\textbf{Ablation study on direct sound optimization}. Ours(w/o opt) is a model that does not have direct sound optimization but includes attentions. Text is ground-truth audio caption of audio datasets \cite{kim2019audiocaps}.} 
\label{figure_ablation_02}
\vspace{-0.5cm}
\end{figure}

\textbf{Ablation: Direct Sound Optimization}
To analyze the impact of Direct Sound Optimization (DSO) on the performance of our model, we conduct an ablation study by constructing a model (w/o opt) that excludes DSO and evaluating its performance on datasets. Table~\ref{table:ablation} shows that the model without DSO lower scores than our model on datasets except CLIPscore on the Audiocaps. In addition, as shown in Table~\ref{table:ablation_human}, in the human evaluation, more than half of the participants answer that the images generated by our model are more suitable for the sound than those generated by the model without DSO. As shown in Figure~\ref{figure_ablation_02}, when given the sound ``Cats meowing loudly and then a male voice", both our model and the model without DSO generate images of a cat and a male, but our model generates an image that is more optimized for the sound and has higher quality in the details compared to the w/o opt model. These results demonstrate that DSO allows the latent $z$ to better optimize the generated images for the sound and improves the quality of the detailed parts in the images.

\begin{figure*}[t] \centering
\includegraphics[width=1\textwidth]{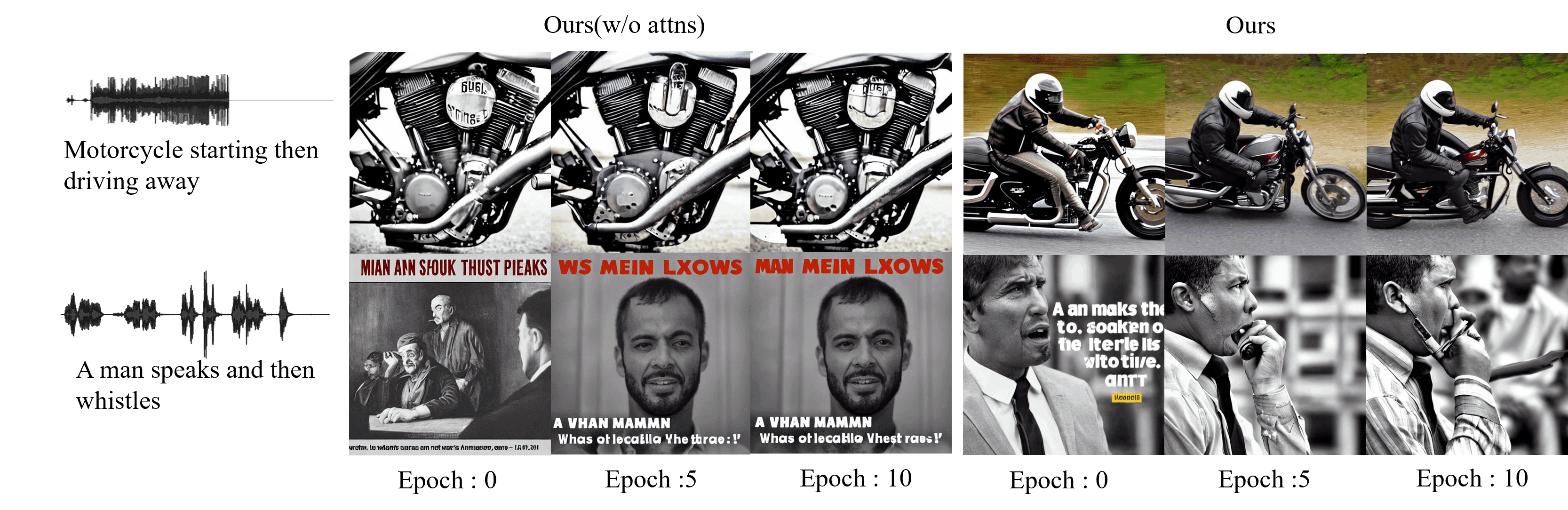}
    \caption{\textbf{Ablation study on all attentions.} Ours(w/o attns) is a model that does not have attentions but includes direct sound optimization. Text is ground-truth audio caption of audio datasets \cite{kim2019audiocaps}.} \label{figure_directsoundoptimiation}
\end{figure*}

\textbf{Ablation: All Attentions}
To analyze the impact of the attentions on the performance of our model, we conduct an ablation study by constructing a model (w/o attns) that excludes attentions and evaluating its performance on all datasets. As shown in Table~\ref{table:ablation}, the model without attentions shows lower scores compared to our model. Furthermore, we conduct a human evaluation to compare the images generated by the model without attentions and our model. More than half of the participants answer that the images generated by our model were more suitable for the given sound compared to the images generated by the model without attentions. Additionally, one might have a concern that Direct Sound Optimization (DSO) could potentially eliminate the functionality of attentions by optimizing the images for the sound. However, as shown in Figure~\ref{figure_directsoundoptimiation}, even with DSO, we observe that the details of the image changes, but the main content of the image still remains. For example, given sound of ``motorcycle starting then driving away", our model generates motorcycles and people, but the w/o attns model generates engines. Even after going through the DSO process, the main contents of the image, such as motorcycles and engines, maintain and only the details will be changed. Therefore, we observe that initializing the $W_{init}$ vector via attentions is very important in representing wild sound. Figure~\ref{figure_optimization} shows the detailed process. 

\begin{figure}[t] \centering
\includegraphics[width=0.5\textwidth]{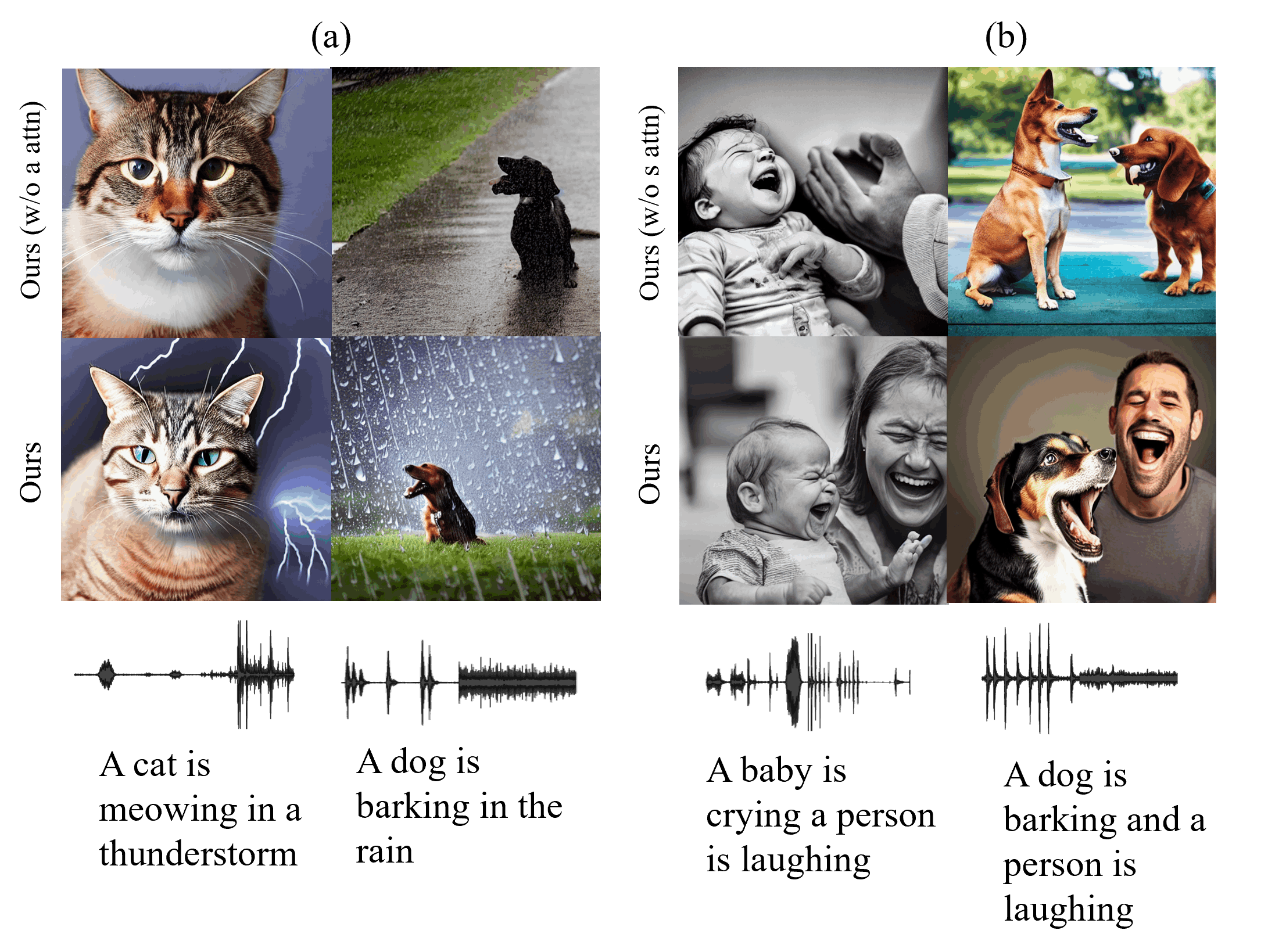}
    \caption{\textbf{Ablation studies on each attention}. Ours(w/o a attn) is a model without audio attention. Ours(w/o s attn) is a model without sentence attention. } 
    \vspace{-0.3cm}
\label{figure_ablation_sa}
\end{figure}

\textbf{Ablation: Audio Attention}
To further analyze the impact of audio attention, we conducted an ablation study on a model without the audio attention (w/o a attn). We use the value of probability, used by the decoder of the ACT model to generate the audio caption as audio attention. It allows for the representation of the dynamic characteristics and rich intensity of wild sounds. According to Table~\ref{table:ablation}, our model slightly outperformed the w/o a attn model on Audiocaps and showed a 5\% higher Yolo score on the Multi-ESC50. Furthermore, as shown in Table~\ref{table:ablation_human}, 59.20\% of users in the human evaluation preferred our model over the w/o a attn model on Audiocaps. As shown in Figure~\ref{figure_ablation_sa} (a), our model can emphasize and visually represent audio features such as intensity of thunder or rain sounds. However, the w/o a attn model without audio attention, sometimes fails to adequately represent these rich audio characteristics.
These results indicate that audio attention sufficiently represents the various characteristics of sound.

\textbf{Ablation: Sentence Attention}
We employed sentence attention to represent objects contained within the audio as images. To analyze the impact of sentence attention, we compared our model to a model without sentence attention (w/o s attn). According to Table~\ref{table:ablation}, our model outperformed w/o s attn model on all datasets. In the Yolo score of Multi-ESC50, our model scored 78.57\%, which is approximately 5\% higher than w/o s attn model score of 72.85\%. Moreover, 58.30\% of participants in the human evaluation preferred our model over the w/o s attn model. As seen in Figure~\ref{figure_ablation_sa} (b), our model can accurately generate objects associated with the sound when given a multi-class sound as input. However, the w/o s attn model sometimes failed to generate accurate objects present in the multi-object sounds. This result indicates that sentence attention assists in generating appropriate objects in images from sounds.

\begin{figure}[t] 
\begin{center}
\centering
\includegraphics[width=0.95\linewidth]{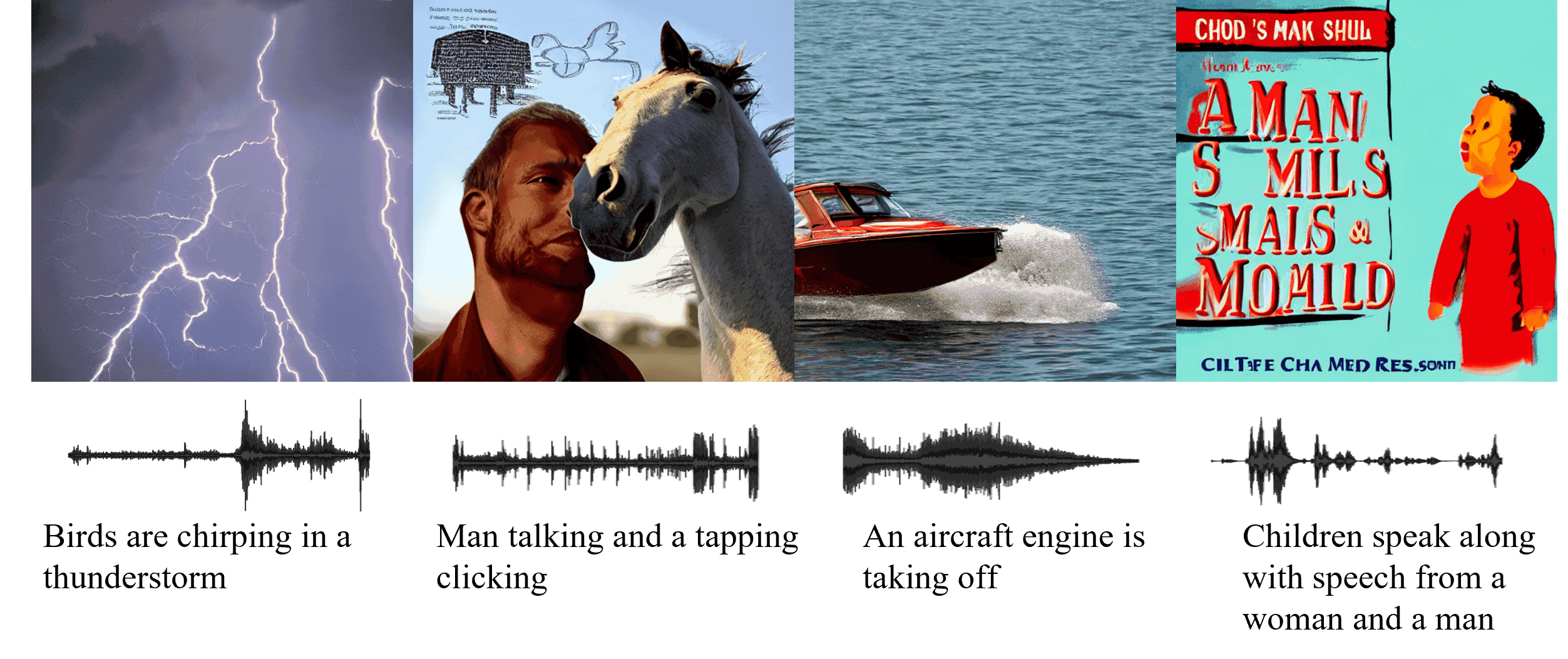}
\end{center}
\caption{\textbf{Failure cases of image generation with our model.}}
\label{figure_exp06}
\vspace{-0.5cm}
\end{figure}

Through these ablation studies, we analyze the effects of the attentions and DSO on our model and observe that using two components better represents wild sounds as images and overcomes the difference in modalities.

\section{Conclusion and Limitation}
We propose a novel approach that uses audio captioning and diffusion-based text-guided image generation model \cite{rombach2022high} to generate a high quality image with open domain from wild sounds. We propose audio attention and sentence attention to represent the dynamic properties of sounds in the wild and leverage CLIPscore\cite{hessel2021clipscore} and AudioCLIP\cite{guzhov2022audioclip} similarities to optimize generated images. As a result, our model successfully generates faithful and high-quality images from wild sound, and outperforms baselines in both quantitative and qualitative evaluations. Nevertheless, our model depends on audio captioning \cite{kim2019audiocaps} and text-guided image generation model \cite{rombach2022high} so we could find some cases where our model fails, such as Figure \ref{figure_exp06}. However, we need neither large paired datasets between sounds and images nor training of the model, and we can still generate high-quality images of open domain from wild sounds.

\section*{Acknowledgement}
 This work was partly supported by Institute of Information \& communications Technology Planning \& Evaluation (IITP) grant funded by the Korea government (MSIT) (No.2022-0-00608, Artificial intelligence research about multi-modal interactions for empathetic conversations with humans \& No.2020-0-01336, Artificial Intelligence
 Graduate School Program (UNIST)) and the Settlement Research Fund (1.210147.01) of UNIST (Ulsan National Institute of Science \& Technology).

{\small
\bibliographystyle{ieee_fullname}
\bibliography{egbib}

\begin{thebibliography}{10}\itemsep=-1pt

\bibitem{akbik2018coling}
Alan Akbik, Duncan Blythe, and Roland Vollgraf.
\newblock Contextual string embeddings for sequence labeling.
\newblock In {\em {COLING} 2018, 27th International Conference on Computational
  Linguistics}, pages 1638--1649, 2018.

\bibitem{aytar2016soundnet}
Yusuf Aytar, Carl Vondrick, and Antonio Torralba.
\newblock Soundnet: Learning sound representations from unlabeled video.
\newblock {\em Advances in neural information processing systems}, 29, 2016.

\bibitem{azar2007sound}
Jimmy Azar, Hassan Abou~Saleh, and Mohamad Al-Alaoui.
\newblock Sound visualization for the hearing impaired.
\newblock {\em International Journal of Emerging Technologies in Learning
  (iJET)}, 2(1), 2007.

\bibitem{child2019generating}
Rewon Child, Scott Gray, Alec Radford, and Ilya Sutskever.
\newblock Generating long sequences with sparse transformers.
\newblock {\em arXiv preprint arXiv:1904.10509}, 2019.

\bibitem{crowson2022vqgan}
Katherine Crowson, Stella Biderman, Daniel Kornis, Dashiell Stander, Eric
  Hallahan, Louis Castricato, and Edward Raff.
\newblock Vqgan-clip: Open domain image generation and editing with natural
  language guidance.
\newblock In {\em European Conference on Computer Vision}, pages 88--105.
  Springer, 2022.

\bibitem{deng2009imagenet}
Jia Deng, Wei Dong, Richard Socher, Li-Jia Li, Kai Li, and Li Fei-Fei.
\newblock Imagenet: A large-scale hierarchical image database.
\newblock In {\em 2009 IEEE conference on computer vision and pattern
  recognition}, pages 248--255. Ieee, 2009.

\bibitem{dosovitskiy2020image}
Alexey Dosovitskiy, Lucas Beyer, Alexander Kolesnikov, Dirk Weissenborn,
  Xiaohua Zhai, Thomas Unterthiner, Mostafa Dehghani, Matthias Minderer, Georg
  Heigold, Sylvain Gelly, et~al.
\newblock An image is worth 16x16 words: Transformers for image recognition at
  scale.
\newblock {\em arXiv preprint arXiv:2010.11929}, 2020.

\bibitem{drossos2020clotho}
Konstantinos Drossos, Samuel Lipping, and Tuomas Virtanen.
\newblock Clotho: An audio captioning dataset.
\newblock In {\em ICASSP 2020-2020 IEEE International Conference on Acoustics,
  Speech and Signal Processing (ICASSP)}, pages 736--740. IEEE, 2020.

\bibitem{esser2021taming}
Patrick Esser, Robin Rombach, and Bjorn Ommer.
\newblock Taming transformers for high-resolution image synthesis.
\newblock In {\em Proceedings of the IEEE/CVF conference on computer vision and
  pattern recognition}, pages 12873--12883, 2021.

\bibitem{gemmeke2017audio}
Jort~F Gemmeke, Daniel~PW Ellis, Dylan Freedman, Aren Jansen, Wade Lawrence,
  R~Channing Moore, Manoj Plakal, and Marvin Ritter.
\newblock Audio set: An ontology and human-labeled dataset for audio events.
\newblock In {\em 2017 IEEE international conference on acoustics, speech and
  signal processing (ICASSP)}, pages 776--780. IEEE, 2017.

\bibitem{guzhov2022audioclip}
Andrey Guzhov, Federico Raue, J{\"o}rn Hees, and Andreas Dengel.
\newblock Audioclip: Extending clip to image, text and audio.
\newblock In {\em ICASSP 2022-2022 IEEE International Conference on Acoustics,
  Speech and Signal Processing (ICASSP)}, pages 976--980. IEEE, 2022.

\bibitem{hao2022attention}
Wangli Hao, Meng Han, Shancang Li, and Fuzhong Li.
\newblock An attention enhanced cross-modal image--sound mutual generation
  model for birds.
\newblock {\em The Computer Journal}, 65(2):410--422, 2022.

\bibitem{hessel2021clipscore}
Jack Hessel, Ari Holtzman, Maxwell Forbes, Ronan~Le Bras, and Yejin Choi.
\newblock Clipscore: A reference-free evaluation metric for image captioning.
\newblock {\em arXiv preprint arXiv:2104.08718}, 2021.

\bibitem{ho2020denoising}
Jonathan Ho, Ajay Jain, and Pieter Abbeel.
\newblock Denoising diffusion probabilistic models.
\newblock {\em Advances in Neural Information Processing Systems},
  33:6840--6851, 2020.

\bibitem{jeong2021traumerai}
Dasaem Jeong, Seungheon Doh, and Taegyun Kwon.
\newblock Tr{\"a}umerai: Dreaming music with stylegan.
\newblock {\em arXiv preprint arXiv:2102.04680}, 2(4):10, 2021.

\bibitem{karras2019style}
Tero Karras, Samuli Laine, and Timo Aila.
\newblock A style-based generator architecture for generative adversarial
  networks.
\newblock In {\em Proceedings of the IEEE/CVF conference on computer vision and
  pattern recognition}, pages 4401--4410, 2019.

\bibitem{kim2019audiocaps}
Chris~Dongjoo Kim, Byeongchang Kim, Hyunmin Lee, and Gunhee Kim.
\newblock Audiocaps: Generating captions for audios in the wild.
\newblock In {\em Proceedings of the 2019 Conference of the North American
  Chapter of the Association for Computational Linguistics: Human Language
  Technologies, Volume 1 (Long and Short Papers)}, pages 119--132, 2019.

\bibitem{kim2022verse}
Taehoon Kim, Gwangmo Song, Sihaeng Lee, Sangyun Kim, Yewon Seo, Soonyoung Lee,
  Seung~Hwan Kim, Honglak Lee, and Kyunghoon Bae.
\newblock L-verse: Bidirectional generation between image and text.
\newblock In {\em Proceedings of the IEEE/CVF Conference on Computer Vision and
  Pattern Recognition}, pages 16526--16536, 2022.

\bibitem{kingma2014adam}
Diederik~P Kingma and Jimmy Ba.
\newblock Adam: A method for stochastic optimization.
\newblock {\em arXiv preprint arXiv:1412.6980}, 2014.

\bibitem{lee2020crossing}
Cheng-Che Lee, Wan-Yi Lin, Yen-Ting Shih, Pei-Yi Kuo, and Li Su.
\newblock Crossing you in style: Cross-modal style transfer from music to
  visual arts.
\newblock In {\em Proceedings of the 28th ACM International Conference on
  Multimedia}, pages 3219--3227, 2020.

\bibitem{lee2022sound}
Seung~Hyun Lee, Wonseok Roh, Wonmin Byeon, Sang~Ho Yoon, Chanyoung Kim, Jinkyu
  Kim, and Sangpil Kim.
\newblock Sound-guided semantic image manipulation.
\newblock In {\em Proceedings of the IEEE/CVF Conference on Computer Vision and
  Pattern Recognition}, pages 3377--3386, 2022.

\bibitem{li2018creating}
Bochen Li, Xinzhao Liu, Karthik Dinesh, Zhiyao Duan, and Gaurav Sharma.
\newblock Creating a multitrack classical music performance dataset for
  multimodal music analysis: Challenges, insights, and applications.
\newblock {\em IEEE Transactions on Multimedia}, 21(2):522--535, 2018.

\bibitem{mei2021audio}
Xinhao Mei, Xubo Liu, Qiushi Huang, Mark~D Plumbley, and Wenwu Wang.
\newblock Audio captioning transformer.
\newblock {\em arXiv preprint arXiv:2107.09817}, 2021.

\bibitem{nichol2021glide}
Alex Nichol, Prafulla Dhariwal, Aditya Ramesh, Pranav Shyam, Pamela Mishkin,
  Bob McGrew, Ilya Sutskever, and Mark Chen.
\newblock Glide: Towards photorealistic image generation and editing with
  text-guided diffusion models.
\newblock {\em arXiv preprint arXiv:2112.10741}, 2021.

\bibitem{patashnik2021styleclip}
Or Patashnik, Zongze Wu, Eli Shechtman, Daniel Cohen-Or, and Dani Lischinski.
\newblock Styleclip: Text-driven manipulation of stylegan imagery.
\newblock In {\em Proceedings of the IEEE/CVF International Conference on
  Computer Vision}, pages 2085--2094, 2021.

\bibitem{piczak2015esc}
Karol~J Piczak.
\newblock Esc: Dataset for environmental sound classification.
\newblock In {\em Proceedings of the 23rd ACM international conference on
  Multimedia}, pages 1015--1018, 2015.

\bibitem{radford2021learning}
Alec Radford, Jong~Wook Kim, Chris Hallacy, Aditya Ramesh, Gabriel Goh,
  Sandhini Agarwal, Girish Sastry, Amanda Askell, Pamela Mishkin, Jack Clark,
  et~al.
\newblock Learning transferable visual models from natural language
  supervision.
\newblock In {\em International Conference on Machine Learning}, pages
  8748--8763. PMLR, 2021.

\bibitem{raffel2020exploring}
Colin Raffel, Noam Shazeer, Adam Roberts, Katherine Lee, Sharan Narang, Michael
  Matena, Yanqi Zhou, Wei Li, Peter~J Liu, et~al.
\newblock Exploring the limits of transfer learning with a unified text-to-text
  transformer.
\newblock {\em J. Mach. Learn. Res.}, 21(140):1--67, 2020.

\bibitem{ramesh2022hierarchical}
Aditya Ramesh, Prafulla Dhariwal, Alex Nichol, Casey Chu, and Mark Chen.
\newblock Hierarchical text-conditional image generation with clip latents.
\newblock {\em arXiv preprint arXiv:2204.06125}, 2022.

\bibitem{ramesh2021zero}
Aditya Ramesh, Mikhail Pavlov, Gabriel Goh, Scott Gray, Chelsea Voss, Alec
  Radford, Mark Chen, and Ilya Sutskever.
\newblock Zero-shot text-to-image generation.
\newblock In {\em International Conference on Machine Learning}, pages
  8821--8831. PMLR, 2021.

\bibitem{redmon2016you}
Joseph Redmon, Santosh Divvala, Ross Girshick, and Ali Farhadi.
\newblock You only look once: Unified, real-time object detection.
\newblock In {\em Proceedings of the IEEE conference on computer vision and
  pattern recognition}, pages 779--788, 2016.

\bibitem{rombach2022high}
Robin Rombach, Andreas Blattmann, Dominik Lorenz, Patrick Esser, and Bj{\"o}rn
  Ommer.
\newblock High-resolution image synthesis with latent diffusion models.
\newblock In {\em Proceedings of the IEEE/CVF Conference on Computer Vision and
  Pattern Recognition}, pages 10684--10695, 2022.

\bibitem{saharia2022photorealistic}
Chitwan Saharia, William Chan, Saurabh Saxena, Lala Li, Jay Whang, Emily
  Denton, Seyed Kamyar~Seyed Ghasemipour, Burcu~Karagol Ayan, S~Sara Mahdavi,
  Rapha~Gontijo Lopes, et~al.
\newblock Photorealistic text-to-image diffusion models with deep language
  understanding.
\newblock {\em arXiv preprint arXiv:2205.11487}, 2022.

\bibitem{salamon2014dataset}
Justin Salamon, Christopher Jacoby, and Juan~Pablo Bello.
\newblock A dataset and taxonomy for urban sound research.
\newblock In {\em Proceedings of the 22nd ACM international conference on
  Multimedia}, pages 1041--1044, 2014.

\bibitem{salimans2016improved}
Tim Salimans, Ian Goodfellow, Wojciech Zaremba, Vicki Cheung, Alec Radford, and
  Xi Chen.
\newblock Improved techniques for training gans.
\newblock {\em Advances in neural information processing systems}, 29, 2016.

\bibitem{schuhmann2022laion}
Christoph Schuhmann, Romain Beaumont, Richard Vencu, Cade Gordon, Ross
  Wightman, Mehdi Cherti, Theo Coombes, Aarush Katta, Clayton Mullis, Mitchell
  Wortsman, et~al.
\newblock Laion-5b: An open large-scale dataset for training next generation
  image-text models.
\newblock {\em arXiv preprint arXiv:2210.08402}, 2022.

\bibitem{shim2021s2i}
Joo~Yong Shim, Joongheon Kim, and Jong-Kook Kim.
\newblock S2i-bird: Sound-to-image generation of bird species using generative
  adversarial networks.
\newblock In {\em 2020 25th International Conference on Pattern Recognition
  (ICPR)}, pages 2226--2232. IEEE, 2021.

\bibitem{touvron2021training}
Hugo Touvron, Matthieu Cord, Matthijs Douze, Francisco Massa, Alexandre
  Sablayrolles, and Herv{\'e} J{\'e}gou.
\newblock Training data-efficient image transformers \& distillation through
  attention.
\newblock In {\em International Conference on Machine Learning}, pages
  10347--10357. PMLR, 2021.

\bibitem{van2017neural}
Aaron Van Den~Oord, Oriol Vinyals, et~al.
\newblock Neural discrete representation learning.
\newblock {\em Advances in neural information processing systems}, 30, 2017.

\bibitem{vaswani2017attention}
Ashish Vaswani, Noam Shazeer, Niki Parmar, Jakob Uszkoreit, Llion Jones,
  Aidan~N Gomez, {\L}ukasz Kaiser, and Illia Polosukhin.
\newblock Attention is all you need.
\newblock {\em Advances in neural information processing systems}, 30, 2017.

\bibitem{viazovetskyi2020stylegan2}
Yuri Viazovetskyi, Vladimir Ivashkin, and Evgeny Kashin.
\newblock Stylegan2 distillation for feed-forward image manipulation.
\newblock In {\em Computer Vision--ECCV 2020: 16th European Conference,
  Glasgow, UK, August 23--28, 2020, Proceedings, Part XXII 16}, pages 170--186.
  Springer, 2020.

\bibitem{wan2019towards}
Chia-Hung Wan, Shun-Po Chuang, and Hung-Yi Lee.
\newblock Towards audio to scene image synthesis using generative adversarial
  network.
\newblock In {\em ICASSP 2019-2019 IEEE International Conference on Acoustics,
  Speech and Signal Processing (ICASSP)}, pages 496--500. IEEE, 2019.

\bibitem{wu2022wav2clip}
Ho-Hsiang Wu, Prem Seetharaman, Kundan Kumar, and Juan~Pablo Bello.
\newblock Wav2clip: Learning robust audio representations from clip.
\newblock In {\em ICASSP 2022-2022 IEEE International Conference on Acoustics,
  Speech and Signal Processing (ICASSP)}, pages 4563--4567. IEEE, 2022.

\bibitem{yang2020diverse}
Pei-Tse Yang, Feng-Guang Su, and Yu-Chiang~Frank Wang.
\newblock Diverse audio-to-image generation via semantics and feature
  consistency.
\newblock In {\em 2020 Asia-Pacific Signal and Information Processing
  Association Annual Summit and Conference (APSIPA ASC)}, pages 1188--1192.
  IEEE, 2020.

\bibitem{zhang2018study}
Yingfang Zhang, Yi Pan, and Junren Zhou.
\newblock Study on application of audio visualization in new media art.
\newblock In {\em Journal of Physics: Conference Series}, volume 1098, page
  012003. IOP Publishing, 2018.

\end{thebibliography}
}

\end{document}